# Conservation AI: Live Stream Analysis for the Detection of Endangered Species Using Convolutional Neural Networks and Drone Technology

C. Chalmers, P.Fergus, Serge Wich and Aday Curbelo Montanez

*Abstract*— **Many different species are adversely affected by poaching. In response to this escalating crisis, efforts to stop poaching using hidden cameras, drones and DNA tracking have been implemented with varying degrees of success. Limited resources, costs and logistical limitations are often the cause of most unsuccessful poaching interventions. The study presented in this paper outlines a flexible and interoperable framework for the automatic detection of animals and poaching activity to facilitate early intervention practices. Using a robust deep learning pipeline, a convolutional neural network is trained and implemented to detect rhinos and cars (considered an important tool in poaching for fast access and artefact transportation in natural habitats) in the study, that are found within live video streamed from drones Transfer learning with the Faster RCNN Resnet 101 is performed to train a custom model with 350 images of rhinos and 350 images of cars. Inference is performed using a frame sampling technique to address the required trade-off control precision and processing speed and maintain synchronisation with the live feed. Inference models are hosted on a web platform using flask web serving, OpenCV and TensorFlow 1.13. Video streams are transmitted from a DJI Mavic Pro 2 drone using the Real-Time Messaging Protocol (RMTP). The best trained Faster RCNN model achieved a mAP of 0.83 @IOU 0.50 and 0.69 @IOU 0.75 respectively. In comparison an SSD-mobilenetmodel trained under the same experimental conditions achieved a mAP of 0.55 @IOU .50 and 0.27 @IOU 0.75.The results demonstrate that using a FRCNN and off-the-shelf drones is a promising and scalable option for a range of conservation projects.**

## I. Introduction

Poaching is a global issue that impacts many animal species [1]. For instance, none more so than the Rhinocerotidae (rhinoceros or rhino for short) where approximately 6000 rhinoceroses have been illegally hunted in South Africa since 2008[1]. The current level of poaching and their subsequent loss as a result of poaching activity has left rhinoceros species at real risk of becoming extinct in the near future [2]. The driving factor behind rhino poaching is the illegal trading of rhino horns which has increased exponentially from $4700 per kilogram in 1993 to $65,000 per kilogram in 2012 [3]. Demand for rhino horn is largely driven by the belief that it can be used in medicinal treatments and as a status symbol within many cultures. The market is mainly focused in Asia, particularly China and Vietnam, which has a growing consumer base. Existing solutions for preventing rhino poaching, have unfortunately shown marginal effect and so rhino poaching continues. For example, GPS trackers have been attached to animals to track and locate animals in natural habitats – when a loss of GPS movement is detected suspected occurrences of poaching is inferred and later confirmed when the injured or dead animal is found [4]. This is a reactive approach that does not prevent the poaching event [5]. Preventive efforts are in place to detect poachers using drones [6], [7], including AirShepherd[2]. However, they either required in-field human analysis or off-line processing of video footage. Therefore, the goal is to use computer vision to automate the detection process and identify objects of interest (Lambda et al. 2019).

The use of computer vision has increased dramatically, largely due to advancements in deep learning (DL), high performance hardware and data availability. Examples of computer vision applications include those in medicine [8], manufacturing [9], and engineering [10]. DL in particular, has facilitated developments in image processing, none more so that through the availability of frameworks like TensorFlow [11], Microsoft Cognitive Tool Kit (CNTK) [12] and Caffe 2 [13]. Convolutional Neural Networks (CNNs) have become the gold standard for image processing tasks, in particular, object detection and segmentation, which are now considered important techniques for analysing animals in video footage. The computational requirements to train and host CNNs, such as the Faster-Region-based Convolutional Neural Network (Faster-RCNN), is a significant challenge, particularly when real-time video processing is required, due to their complex network architectures. Graphical processing units (GPUs) help, however enterprise-level cards are expensive and trade-offs are required depending on the application and services required.

Nonetheless, object detection has been incorporated into numerous conservation projects [14]. Solutions are typically focused on low-resolution images and lightweight models such as You Only Look Once (YOLO) [15] and the Single Shot

---

[1] https://www.environment.gov.za/mediarelease/molewa_highlightsprogress_againstrhinopoaching

[2] https://airshepherd.org/

Multibox Detector (SSD)-mobilenet [16]. While they have a useful place in conservation, they have limited utility when detailed image analysis is required. RCNNs incorporate region-based proposals which significantly improves the accuracy of object detections overall but particularly in noisy images affected by distance, terrain and occlusion [17]. This is often confounded with the fact analyses is conducted offline by domain experts.

In this study we propose an alternative approach that combines off-the-shelf drones with near-real time detection of rhinos and cars for different conservation tasks.

The remainder of the paper is structured as follows. A background discussion on current anti-poaching solutions and their associated limitations is introduced in Section 2. Section 3 details the proposed methodology before the results are presented in Section 4. Section 5 discusses the results before the paper is concluded and future work is presented in Section 6.

## II. RELATED WORK

Wildlife management systems have become increasingly popular in poaching prevention strategies. Sophisticated, efficient and cost-effective solutions exist yet there is still little evidence to show that poaching is not being effectively controlled [18], [19]. The remainder of this section provides a discussion on some of the current systems in operation today and highlights the key limitations in each approach that this paper aims to address.

### A. Current Solutions

Camera systems are used extensively in conservation to observe wildlife and detect poaching activity [20]. The identification of objects, such as animals, in video footage has typically been undertaken by humans. However, there is now significant interest in automating this process using computer vision technology. While significant progress has been made in model accuracy, inferencing speed has become the focal point within the conservation domain. For example, in [6] a Faster-RCNN was implemented to identify poachers and animals using videos obtained from drones. The results for video processing with CPU, GPU and cloud processing are presented to benchmark inferencing time. The key findings show that a Faster-RCNN model inferences at approximately 5 frames per second (fps) using a single K40 GPU. Live video stream on the other hand runs at 25 (fps). This creates a bottleneck in the detection pipeline and suffers with issues surrounding video synchronisation. Similar results were reported in [21] using the ImageNet dataset. However, the low-resolution images used are unsuitable for natural environments and conservation tasks due to the inability to capture the necessary features for both training and inferencing.

Faster-RCNN are powerful models in object detection, but meeting the computational requirements needed for training and near real-time inference is costly. This is why many conservation strategies utilise lightweight models, such as YOLO and SSD Mobilenet, as they are much easier to train and inference on. For example, [22] use a YOLO model install on a Nvidia Jetson TX1 attached to a drone is trained to detect pedestrians. The results show it was possible to inference 2 fps with an image input size of 256 * 256 pixels. In practice such systems would provide limited utility as higher altitudes in conservation would be required for safety and conservational needs. In a similar approach [23] implemented an optimal dense YOLO method to detect vehicles at greater altitudes. The drone was flown between 60-70 metres off the ground and using YOLOv3 the model could reach 86.7% mAP with a frame rate of 17 fps compared with YOLOv2, which achieved a 49.3% mAP and a frame rate of 52 fps. The paper does not include the IOU metrics making it difficult to evaluate how well the model performed.

An ensemble model is presented in [24] comprising a lightweight image region detector to identify objects of interest. Each identified frame is processed in a cloud infrastructure to classify objects. The local node decides if the current frame contains an object of interest using the Binarized Normed Gradients (BING) algorithm to measure the objectness on every input frame. An image resolution of 320 x 240 pixels at 60 (fps) was used for the initial detection. The paper reports a reasonable inference time of 1.29 seconds using a cloud hosted Faster-RCNN. However, the results were only possible when the drone was flown a few metres above the object of interest. Again, for safety and conservational needs this approach would find limited utility in complex conservational projects.

### B. Limitations

The studies previously discussed combine drones, computer vision and machine learning to provide new and interesting tools for use in conservational studies. While, they all provide some interesting results, several challenges still remain. In particular, studies that utilise lightweight models show that while inference time is fast the overall accuracy of objects detected is poor. This is a particular challenge in conservation when flying at higher altitudes is necessary to ensure that animals are not disturbed and poachers are not alerted. It has been widely reported that YOLO suffers from spatial constraints and this consequently limits the number of nearby objects that can be detected [25]. In other words YOLO-based models perform poorly when detecting small objects or objects that appear in groups or which are occluded. YOLO down samples images to 448 * 448 pixels and lower. So, while it is possible to inference in excess of 45(fps) using a Titan X GPU [25] and support real-time inferencing, down sampling results in a significant loss of attainable features, which in turn decreases the overall accuracy of the model.

Networks, such as Faster-RCNN, are now being used to address many of these issues through region-based proposals that act like sliding windows during the feature extraction process. This helps to significantly improve overall detection accuracy, but it dramatically increase inferencing time due to the network and underling pipeline being significantly more complex than YOLO-based solutions [26]. Advancements have been made since the original R-CNN was proposed in terms of speed and efficiency, however, region-based proposal networks are still problematic for real-time inferencing. Integrating region-based models into the detection pipeline has led to a greater reliance on GPU compute. This allows for much faster inference but brings with it additional challenges. It is difficult to implement and use GPU-based solutions in the field due their unique power consumption needs, thus wide area communications is needed to support remove inference, which is not always possible. Furthermore, real-time inference is costly and often requires a trade-off between application-specific requirements and the amount of GPU compute needed. In this paper we investigate this issue further and propose a viable solution for live object detection in natural environments to support complex conservation field trials and application specific requirements.

### III. METHODOLOGY

This section describes the data collection and training process for the object detection model. Training parameters and associated techniques are discussed along with the proposed frame sampling technique required for inferencing and support different application requirements. This is followed by a discussion on the drone and communication protocol used for live video streaming and object detection locally and remote servers. Finally, the evaluation metrics used to assess both the trained model's performance and the frame sampling technique are formalised and discussed.

#### A. Data capture

The dataset comprises RGB and thermal (colour and grey scale palettes) images. The data contains two classes: rhinos and cars. Each class has 350 images with resolutions between 300 x 147 and 3840 x 2160 pixels. To maintain sufficient variance aerial footage is combined with close exposure and thermal images as shown in Figure 1. The aerial RGB images are captured using a drone while both the thermal and grey scale images where captured using a ground-based camera. To supplement our own field trial acquired images, RGB data is batched downloaded from google images.

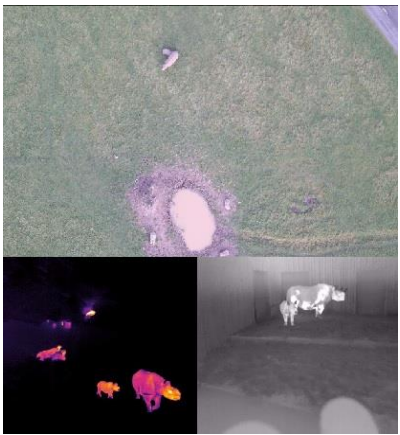

**Fig. 1:** Example training data with variance

---
[3] https://github.com/microsoft/VoTT

Thermal and grey scale images are captured using a FLIR One smart phone thermal camera. The resolution for both the thermal and grey scale images is set to 640 x 512 pixels. These images where collected by Liverpool John Moores University at Knowsley Safari.

Tagging of the data is undertaken using the Visual Object Tagging Tool (VoTT) version 1.7.0[3]. Binding boxes are used to identify regions of interest. All of the tagged regions in each image are exported as Extensible Mark-up Language (XML) in TensorFlow Pascal VOC format. The generated XML is converted into Comma Separated Values (CSV) using the Python libraries Pandas and XML to parse the XML data. Using TensorFlow 1.13.1 and Pillow, both the CSV and image data is converted into the TFRecord format ready for training.

#### B. Model selection Faster-RCNN

The Faster-RCNN network architecture is implemented to perform object detection in two distinct stages [27]. The first stage uses Region Proposal Networks (RPNs) to identify and extract features from the selected layers. This allows the model to estimate binding box locations. The second stage adjusts the localisation of the bounding box by minimising the selected loss function. Both the region proposal and object detection tasks are undertaken by the same CNN. This method offers improvements in terms of speed and accuracy over early R-CNN networks where region proposals were input at the pixel level opposed to the feature map level. The Faster-RCNN gains further speed improvements by replacing selective search with a RPN. Figure 2 shows the basic architecture of a Faster-RCNN.

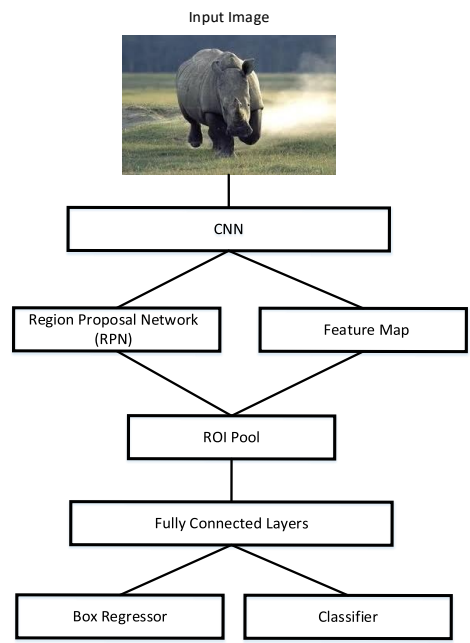

**Fig. 2:** Faster-RCNN Architecture

The restricted nature in which the model must operate (e.g. small or partially occluded objects) means that other non-region-based proposal networks would find it difficult to attain a high degree of accuracy. Additionally, the down sampling undertaken in such models reduces the features available in images.

*C. Transfer learning*

Transfer learning allows us to adopt a pre-trained model (trained on millions of images) and fine-tune the learned parameters during the training process using our rhino and car images [28]. This is an important technique as training CNNs on small datasets leads to extreme overfitting due to low variance. The base model adopted in this study for the transfer learning tasks is the Faster-RCNN Resnet 101 model which is pre-trained using the COCO dataset. COCO is a large object detection dataset containing 330 thousand images and 1.5 million object instances.

*D. Model training*

Model training is conducted on a HP ProLiant ML 350 Gen 9 server. The server has x2 Intel Xeon E5-2640 v4 series processors and 768GB of RAM. An additional GPU stack comprising x4 NVidia Quadro M4000 graphics cards with a combined total of 32GB of DDR5 RAM is installed. TensorFlow 1.13.1, CUDA 10.0 and CuDNN version 7.5 form the software aspects of the training pipeline. In the pipeline.config file used by TensorFlow; the following training parameters are set:

- The aspect ratio resizer minimum and maximum coefficients are set to 1500 x 1500 pixels respectively. This minimises the scaling effect on the acquired data. Increasing the resolution further will facilitate greater accuracy but would hit the computational limitations of the training platform.
- The default for the feature extractor coefficient is retained to provide a standard 16-pixel stride length to maintain a high-resolution aspect ratio.
- The batch size coefficient is set to one to maintain GPU memory limits.

The Resnet 101 model implements the Adam optimiser to minimise the loss function [27]. Unlike other optimisers such as Stochastic Gradient Decent (SGD), which maintains a single learning rate (alpha) throughout the entire training session. Adam calculates the moving average of the gradient $m_t$/squared gradients $v_t$ and the parameters beta1/beta2 to dynamically adjust the learning rate. Adam as described in [28] is defined as:

$$m_t = \beta_1 m_t - 1 + (1 - \beta_1)g_t \quad (1)$$
$$v_t = \beta_2 v_t - 1 + (1 - \beta_2)g_t^2$$

Where $m_t$ and $v_t$ are the estimates of the first and second moment of the gradients. Both $m_t$ and $v_t$ are initialised with 0's. Biases are corrected by computing the first and second moment estimates [28]:

$$\hat{m}_t = \frac{m_t}{1 - \beta_1^t} \quad (2)$$
$$\hat{v}_t = \frac{v_t}{1 - \beta_2^t}$$

Parameters are updated using the Adam update rule:

$$\theta_{t+1} = \theta_t - \frac{n}{\sqrt{\hat{v}_t} + \epsilon} \hat{m}_t. \quad (3)$$

The ReLU activation function during training provides improvements over other functions such as sigmoid or hyperbolic tangent (tanh) activations that suffer from saturation changes around the mid-point of their input. This reduces the amount of available tuning. Their use in deep multi-layered networks result in ineffective training caused by a vanishing gradient [29], ReLU as defined in [30] is:

$$g(x) = max(0, x) \quad (4)$$

For baseline comparison we also train an alternate SSD-mobilenet v2 model. The model is evaluated under the same experimental condition. The pipline.config file is unaltered therefore using the default training parameters:

- The aspect ratio resizer (image_resizer) minimum and maximum coefficients are set to 300 x 300 pixels respectively.

The model is trained over the same number of epochs to ensure consistency in the reported results.

*E. Inferencing pipeline*

The object detection system proposed interfaces with a variety of camera systems using the Real-Time Messaging Protocol (RTMP). The Mavic Pro 2 drone system is used in this study which is capable of transmitting 4K videos at 30fps over a distance in excess of 7 kilometres (km). The done is connected to a linked controller using the OcuSync 2.0 protocol. Video streams at re-directed from the controller using a local Wi-Fi connection to a field laptop or to a remote server using 4G. Object detection on video frames is then performed on the laptop or remote server. Figure 3 illustrates the end-to-end inferencing pipeline.

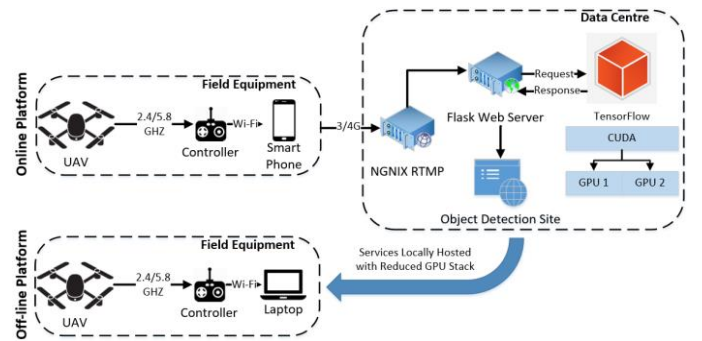

**Fig. 3:** Object Detection Pipeline

The video stream from the controller is transmitted to an RTMP sever hosted by NGNIX. Each frame is processed in Python using OpenCV and serialised with a flask web server. Each video frame is inferenced with a TensorFlow backend model using a loaded saved tensorflow model. The binding box detections are returned to the flask object detection site where the operator can monitor the results. Where local processing is used the complete framework is run locally on a field laptop. In this study local inferencing is performed using a Dell XPS 15 laptop with an Intel i9 CPU and Nvidia 1050Ti GPU. In the case where 4G is used online inferencing is undertaken on a custom-built server containing an Intel Xeon E5-1630v3 CPU, 64GB of RAM and an NVidia Quadro RTX 8000 GPU. TensorFlow 1.13.1, CUDA 10 and CuDNN 7.5 represent the software components used for inferencing. All of the required services and components are available through a public facing conservation AI portal[4] developed by the research team

To overcome the inferencing limitations associated with Faster-RCNNs and the requirement to provide real-time object detection a frame sampling technique is proposed. A single adjustable parameter specifies the number of frames to be inferenced within a given time period. This results in frame skipping which maintains video speed integrity and accurate binding box overlay, i.e. real-time synchronisation with live video streams. This is a configurable parameter that can be adjusted within the object detection site to meet the needs of the underlying GPU architecture. The frame sampling technique is outlined in algorithm 1.

| **Algorithm 1** Dynamic Frame Sampling | |
|---|---|
| **Data:** | Frame Read Correctly *ret*, |
| | Current Video Frame *frame*, |
| | Inferencing Parameter *fpsLimitStream*, |
| | Start Time *startTime*, |
| | Now Time *nowTime* |
| **Result:** | Tracked Bounding Boxes *inference* in *frame* |
| **Read** | *ret, frame* |
| **Set** | *startTime* |
| **While** | *ret* = TRUE |
| | **Set** *nowTime* |
| | **if** *nowTime – startTime > fpsLimitStream* |
| |    **Run** *inference;* |
| |    **Set** *startTime;* |
| | *ret, frame*(next); |

*F. Evaluation metrics*

The model's performance is evaluated using mAP (mean average precision), which is a standard metric for measuring the performance of an object detection model. mAP is defined as:

$$mAP = \frac{\sum_{q=1}^{Q} AveP(q)}{Q} \qquad (5)$$

Where $Q$ is the number of queries in the set and $AveP(q)$ is the average precision ($AP$) for a given query $q$.

The mAP is calculated on the binding box locations for the final two checkpoints. Two Intersection over Union (IOU) thresholds, 0.50 and 0.75, are used to assess the overall performance of the model. The IOU is a useful metric for determining the accuracy of the binding box location. This is achieved by measuring the percentage ratio of the overlap between the predicted bounding box and the ground truth bounding box [24]. A threshold of 0.50 measures the overall detection accuracy while the upper threshold of 0.75 measures localisation accuracy.

To assess the performance of the frame sampling technique and determine the optimal configuration for the underlying GPU the following processing metrics are used:

- **Decode Setting (DC)** describes the total number of frames to be analysed within a specified time period. The coefficient value can be set between 1 and 0.0001 therefore controlling the number of frames to be inferenced. The higher end range reduces the number of frames serialised for inferencing, which increases playback speed. The lower end increases the number of frames to inference therefore decreasing the playback speed. The trade-off between the two metrics is application specific.
- **Video Frame Rate** specifies the frequency rate of the consecutively captured frames from the video source.
- **Total Video Frames** provides the total amount of frames transmitted from the feed based on current playback time.
- **Total Video Frames Analysed (TVFA)** is the number of frames processed for inferencing within the total duration of the video. This metric is calculated in algorithm 1 by counting each of the frames submitted to the object detection model.
- **Percentage of Frames Analysed (PFA)** is the TVFA x 100 / Total Video Frames.
- **Runtime(s)** indicates how long the framework took to process all of the analysed frames (TVFA).
- **Total Video Time(s)** is the total length of the video.
- **Processed Frames per Second (PFPS)** is the Total Video Frames divided by the run time.

IV. EVALUATION

In this section, the performance of the model and the associated frame sampling technique is evaluated using the metrics outlined in the methodology section. The experiments were performed during a field trial at Knowsley Safari in Liverpool in the United Kingdom.

---

[4] www.conservationai.co.uk

## A. Training

In the case of the Faster-RCNN Model training was performed over 3457 epochs to minimise the cost function through backpropagation. The total loss at epoch 3457 was 0.6539, which increased slightly from 0.6395 at epoch 3071 suggesting an overshot of the global minimum. Figure 3 illustrates the total loss for the training session on the y-axis and the epochs on the x-axis.

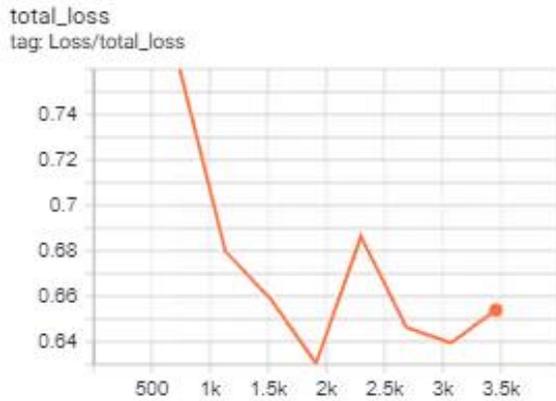

**Fig. 3:** Total Loss Faster-RCNN

In the case of the SSD-mobilenet model training was performed over 4000 epochs to ascertain if the total loss could be reduced. As shown in figure 4 the loss at 4000 epochs was 5.733 while the total loss was 5.439 at epoch 3457 showing minimal improvement.

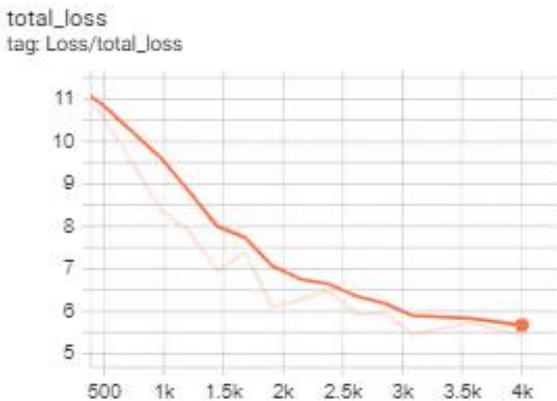

**Fig. 4:** Total Loss SSD-mobilenet

The mAP for the Faster-RCNN model was evaluated for each binding box location for the final two checkpoints (epochs 3071 and 3457) respectively. At epoch 3457, the model attained a mAP @.50IOU of 0.83 and 0.69 @.75IOU. At epoch 3017 the mAP was 0.80 @.50IOU and 0.68 @.75IOU. The performance of the model is shown in figures 5 and 6 where the mAP is on the y-axis and the epochs on the x-axis.

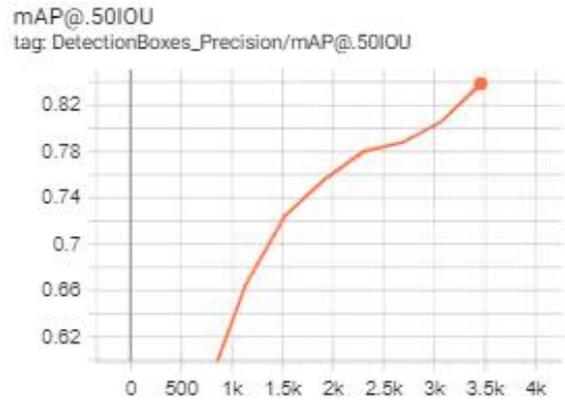

**Fig. 5:** mAP @.50IOU Faster-RCCN

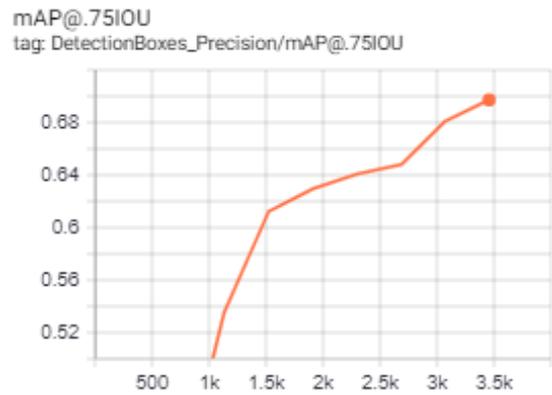

**Fig. 6:** mAP @.75IOU Faster-RCNN

The mAP for the SSD-mobilenet is evaluated at epoch 3457 for direct comparison with the Faster-RCNN model. The model attained a mAP @.50IOU of 0.55 and 0.27 @.75IOU. The performance of the model is highlighted in figures 7 and 5 where the mAP is shown on the y-axis and the epochs on the x-axis.

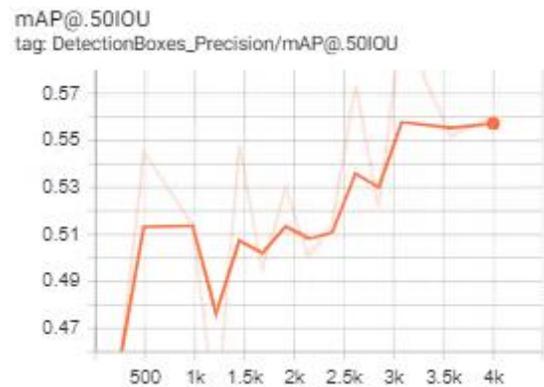

**Fig. 7:** mAP @.50IOU SSD-mobilenet

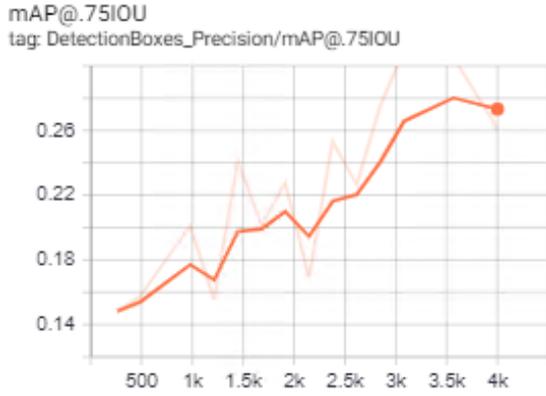

**Fig. 8:** mAP @.75IOU SSD-mobilenet

*B. Inferencing*

To assess the frame sampling technique and to determine the underlying performance of the GPU architecture a test video was uploaded to the object detection pipeline. As shown in figure 9 a sample video classification through the system is demonstrated. The video used to test the framework has a total video time of 0.37 seconds consisting of 925 frames at a resolution of 1280 x 720. The frame rate is set at 25fps. A pre-recorded video was used to baseline the performance of the underlying hardware before real-time inferencing was conducted.

Table I Frame Processing Results

| DC | % FPA | Runtime (s) | PFPS | TVFA |
|---|---|---|---|---|
| **1** | 0.1 | 3 | 308 | 5 |
| **0.1** | 5.9 | 21 | 44 | 55 |
| **0.08** | 7.1 | 29 | 31 | 66 |
| **0.07** | 8.6 | 42 | 22 | 80 |
| **0.06** | 10.07 | 32 | 28.9 | 99 |
| **0.05** | 11.13 | 45 | 20.55 | 103 |
| **0.01** | 52 | 154 | 6 | 482 |
| **0.001** | 53 | 170 | 5.4 | 480 |
| **0.0001** | 52 | 187 | 4.9 | 482 |

Referencing the results in table 1 the optimal coefficients value is 0.05. This provides sufficient frame processing and synchronisation with the live video (i.e. no notable lag is detected). To demonstrate the configuration a field trial was conducted at Knowsley Safari in the UK.

*C. Knowsley Safari Field Trial*

A DJI Mavic Pro 2 was used to stream 4k video at 30fps of the rhino enclosure at Knowsley Safari from an altitude of 60m. Using the decode setting (DC) 0.05 we were able to conduct object detection a 2fps while maintaining synchronisation with the live video stream and ensuring detection boxes remain fixed around objects. The frame sampling technique is the mechanism that allows us to maintain real-time synchronisation with the live video stream whilst conducting live inference. Figure 10 shows the detection of rhinos and cars within the enclosure.

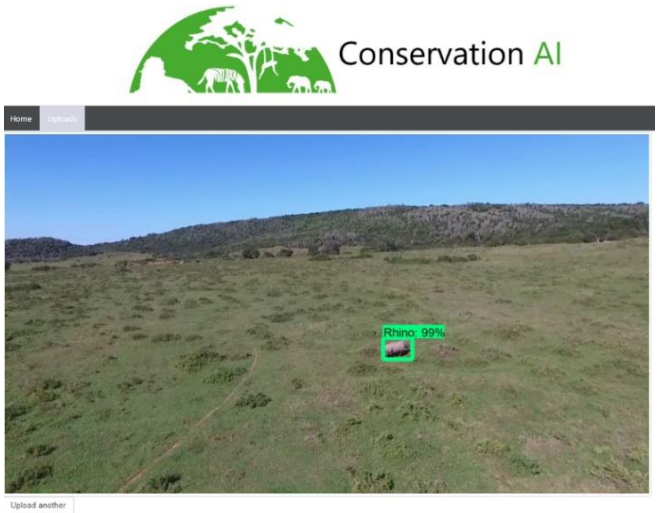

**Fig. 9:** Conservation AI Portal

Using the video configuration previously discussed several experiments where performed to evaluate the precision/speed trade-off and overall video classification performance. These metrics allow us to specify the necessary hardware requirements and overall compute time for live inferencing in production ready systems as well as those for offline video processing tasks. Table 1 presents the results for the frame sampling technique. These results are discussed in more detail later in the paper.

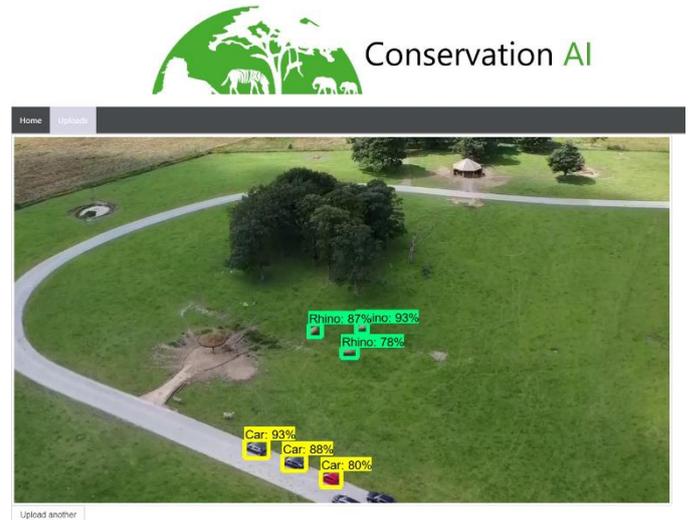

**Fig. 10:** Knowsley Safari Field Trial

In total 11% of the frames collected over the duration of the flight where analysed which equates to approximately 2fps.

Note that the two objects detected within figure 10 include rhinos and cars. Cars were included during training as these are often a key component in poaching activities for fast access and the transportation of rhino horn.

## V. DISCUSSION

In this study we aimed to test an off-the-shelf drone in combination with near-real time object detection to allow for scalability of such solutions in the field. Our system differs from previous solutions such as the bespoke drone in [6], [7].

We discuss our main results here. Firstly, a limited set of 350 images per class where used to train our model. The images in the dataset contained large amounts of variance representing both distance and close up shots as well as thermal colour and grey scale representations. The mAP and IOU are therefore competitive where figures 5 and 6 show that the model was capable of attaining a mAP @.50IOU of 0.83 and 0.69 @.75IOU. Second, the strategic use of transfer learning shows that the detailed features provided in the Faster-RCNN model can be combined with those extracted from our images to detect new kinds of objects (in this case rhinos and cars). Training on the 350 images alone would not support the mAP and IOU results produced in this paper. Lastly using Faster-RCNN and high-quality images without the need to down sample facilities improved detection at higher altitudes as shown in figures 9 and 10 and the corresponding IOU. This overcomes the limitations of approaches reported in this and many studies particularly those that use YOLO and SSD which are widely applied in conservation field trials. The results obtained from our experiments confirm that the use of SSD and similar networks are impractical for many conservation tasks due to both the distance and likelihood of occluded objects.

One of the main novel contributions outlined in this paper is the frameworks ability to maximise precision and throughput while maintaining video detection box synchronisation. This provides a unique platform for integrating systems with different hardware and compute capabilities. These range from high-end compute infrastructures to laptops deployed in the field. This allows the regulation of video frames and provides processing support to a wide variety of camera systems such as drones and camera traps. The results presented in table 1 provide a clear metric between the desired number of frames to analyse verses the compute time needed. For example, a DC of 0.05 facilitates the processing of approximately 11% of captured frames using the trained model. In the example shown in the table 1 a 37-second video containing 925 frames would allow you to process 103 frames over 45 seconds. In contrast, processing 482 (roughly half) would take 187 seconds. The results presented in table 1 where achieved using a NVidia Quadro RTX 8000 GPU costing roughly $10,000. As an example, using industry standard architecture such as a NVidia DGX-1 would cost in excess of $100,000.

The system presented in this paper is capable of working on any infrastructure and will support any necessary compute requirements. Therefore, achieving faster inferencing above and beyond what is shown in this paper will incur increased cost in terms infrastructure and compute requirements in particular the use of enterprise systems such as NVidia DGX1/2 or cloud services. Scaling out to faster computing infrastructure removes the flexibility to conduct onsite inferencing. This said there would be no need to inference faster than 2fps in most conservation projects and we therefore argue that the reported methodology is sufficient.

The feasibility of the system was evaluated in a field trial at Knowsley Safari. Our initial test demonstrated the real-time inferencing capabilities of our approach as shown in figure 7. Due to poor 4G coverage, the DC had to be set at 0.1 which allowed us to process 6% of the total frames acquired. Nonetheless, due to the flexibility of the framework, we were also able to switch to local inferencing using an on-board RTMP server and a laptop containing an NVidia 1050Ti GPU located in the field. This allowed us to set the DC back to 0.05 while maintaining video synchronisation. Note we use the NVidia Quadro RTX 8000 GPU for hosting multiple parallel services which is not possible using none enterprise GPU's, therefore comparable speeds between different GPUs may not be that noticeable.

## VI. CONCLUSIONS AND FUTURE WORK

Many species are adversely affected by poaching and reducing this remains difficult [31]. This paper proposes a novel approach that builds on recent advances in computer vision, machine learning and drone technology. Using advanced neural networks and a transfer learning the system facilitates near real-time detection of rhinos and cars as evidenced in our field trial conducted at Knowlsey Safari in the UK. The framework allows us to personalise the system to the unique properties of the environment. This was demonstrated with the poor 4G coverage experienced in the field trial whereby seamlessly moving to a local solution for inferencing was possible.

The mAP and IOU show that the model was sufficiently trained which is encouraging. Furthermore, the inferencing results highlight the strength of the model in detecting objects in pre-recorded video and live streams. The existing features provided by the frozen graph combined with the features from the dataset using transfer learning significantly improve the classification results. More importantly, the results demonstrate that region-based networks overcome the limitations found with existing approaches such as YOLO and SSD-mobilenet.

The prosed baseline solution will be extended in future work. Concentrated efforts will focus on developing the system further to support complex conservation tasks of which one of the key factors will be scalability. Furthermore, the processing of more complex images is seen as a vital component for better understanding animals and the unique behaviours they exhibit. Therefore, we will also focus research around Neural Architecture Search (NAS) networks, which offer high resolution processing and advanced object detection capabilities. Along with advancements in camera and drone technology, this will allow us to monitor animals at much higher altitudes and this, will help to minimise disruption in natural habitats. In addition, work will be undertaken to include a broader representation of endangered species in object detection tasks.

Several challenges still remain. First, the system outlined in the study relies on a live video stream. This is challenging over large distances. Thus other solutions such as directly uploading images from the drone to a GSM network (if available) to reach a remote server need to be explored. In areas where no GSM network is available and live video feedback is not an option onboard processing of images can be a viable solution as long as results can be transmitted back to rangers on the ground and detections are accurate.

## ACKNOWLEDGMENTS

The authors would like to thank Naomi Davies at Knowsley Safari for co-ordinating the field trial and providing access the rhino enclosure.

## REFERENCES

[1] A. M. Hübschle, "The social economy of rhino poaching: Of economic freedom fighters, professional hunters and marginalized local people," *Curr. Sociol.*, vol. 65, no. 3, pp. 427–447, May 2017.

[2] H. Koen, J. P. de Villiers, H. Roodt, and A. de Waal, "An expert-driven causal model of the rhino poaching problem," *Ecol. Modell.*, vol. 347, pp. 29–39, Mar. 2017.

[3] R. Witter and T. Satterfield, "Rhino poaching and the 'slow violence' of conservation-related resettlement in Mozambique's Limpopo National Park," *Geoforum*, vol. 101, pp. 275–284, 2019.

[4] E. Alphonce Massawe, M. Kisangiri, S. Kaijage, and P. Seshaiyer, "An intelligent real-time wireless sensor network tracking system for monitoring rhinos and elephants in Tanzania national parks: A review," 2017.

[5] T. F. Tan, S. S. Teoh, J. E. Fow, and K. S. Yen, "Embedded human detection system based on thermal and infrared sensors for anti-poaching application," in *2016 IEEE Conference on Systems, Process and Control (ICSPC)*, 2016, pp. 37–42.

[6] E. Bondi *et al.*, "Near Real-Time Detection of Poachers from Drones in AirSim.," in *IJCAI*, 2018, pp. 5814–5816.

[7] E. Bondi *et al.*, "Spot poachers in action: Augmenting conservation drones with automatic detection in near real time," in *Thirty-Second AAAI Conference on Artificial Intelligence*, 2018.

[8] S. Saria, A. Butte, and A. Sheikh, "Better medicine through machine learning: What's real, and what's artificial?" Public Library of Science, 2018.

[9] P. Martinez, R. Ahmad, and M. Al-Hussein, "A vision-based system for pre-inspection of steel frame manufacturing," *Autom. Constr.*, vol. 97, pp. 151–163, 2019.

[10] S. Jakobs, A. Weber, and D. Stapp, "Reliable Object Detection for Autonomous Mobile Machines," *ATZheavy duty Worldw.*, vol. 12, no. 2, pp. 44–49, 2019.

[11] L. Rampasek and A. Goldenberg, "Tensorflow: Biology's gateway to deep learning?," *Cell Syst.*, vol. 2, no. 1, pp. 12–14, 2016.

[12] D. S. Banerjee, K. Hamidouche, and D. K. Panda, "Re-designing CNTK deep learning framework on modern GPU enabled clusters," in *2016 IEEE international conference on cloud computing technology and science (CloudCom)*, 2016, pp. 144–151.

[13] K. Hazelwood *et al.*, "Applied machine learning at facebook: A datacenter infrastructure perspective," in *2018 IEEE International Symposium on High Performance Computer Architecture (HPCA)*, 2018, pp. 620–629.

[14] E. Bondi *et al.*, "Automatic Detection of Poachers and Wildlife with UAVs," *Artif. Intell. Conserv.*, p. 77, 2019.

[15] J. Parham, C. Stewart, J. Crall, D. Rubenstein, J. Holmberg, and T. Berger-Wolf, "An Animal Detection Pipeline for Identification," in *2018 IEEE Winter Conference on Applications of Computer Vision (WACV)*, 2018, pp. 1075–1083.

[16] S.-J. Hong, Y. Han, S.-Y. Kim, A.-Y. Lee, and G. Kim, "Application of Deep-Learning Methods to Bird Detection Using Unmanned Aerial Vehicle Imagery," *Sensors*, vol. 19, no. 7, p. 1651, 2019.

[17] S. Schneider, G. W. Taylor, and S. Kremer, "Deep Learning Object Detection Methods for Ecological Camera Trap Data," in *2018 15th Conference on Computer and Robot Vision (CRV)*, 2018, pp. 321–328.

[18] S. L. Pimm *et al.*, "Emerging technologies to conserve biodiversity," *Trends Ecol. Evol.*, vol. 30, no. 11, pp. 685–696, 2015.

[19] J. Kamminga, E. Ayele, N. Meratnia, and P. Havinga, "Poaching Detection Technologies—A Survey," *Sensors*, vol. 18, no. 5, p. 1474, May 2018.

[20] M. Zeppelzauer and A. S. Stoeger, "Establishing the fundamentals for an elephant early warning and monitoring system," *BMC Res. Notes*, vol. 8, no. 1, p. 409, 2015.

[21] C. L. Zitnick and P. Dollár, "Edge boxes: Locating object proposals from edges," in *European conference on computer vision*, 2014, pp. 391–405.

[22] D. Zhang Sr *et al.*, "Using YOLO-based pedestrian detection for monitoring UAV," in *Tenth International Conference on Graphics and Image Processing (ICGIP 2018)*, 2019, vol. 11069, p. 110693Y.

[23] Z. Xu, H. Shi, N. Li, C. Xiang, and H. Zhou, "Vehicle Detection Under UAV Based on Optimal Dense YOLO Method," in *2018 5th International Conference on Systems and Informatics (ICSAI)*, 2018, pp. 407–411.

[24] J. Lee, J. Wang, D. Crandall, S. Šabanović, and G. Fox, "Real-time, cloud-based object detection for unmanned aerial vehicles," in *2017 First IEEE International Conference on Robotic Computing (IRC)*, 2017, pp. 36–43.

[25] J. Redmon, S. Divvala, R. Girshick, and A. Farhadi, "You only look once: Unified, real-time object detection," in *Proceedings of the IEEE conference on computer vision and pattern recognition*, 2016, pp. 779–788.

[26] S. Ren, K. He, R. Girshick, and J. Sun, "Faster r-cnn: Towards real-time object detection with region proposal networks," in *Advances in neural information processing systems*, 2015, pp. 91–99.

[27] D. P. Kingma and J. Ba, "Adam: A method for stochastic optimization," *arXiv Prepr. arXiv1412.6980*, 2014.

[28] M. Heusel, H. Ramsauer, T. Unterthiner, B. Nessler, and S. Hochreiter, "Gans trained by a two time-scale update rule converge to a local nash equilibrium," in *Advances in Neural Information Processing Systems*, 2017, pp. 6626–6637.

[29] S. Kanai, Y. Fujiwara, and S. Iwamura, "Preventing gradient explosions in gated recurrent units," in *Advances in neural information processing systems*, 2017, pp. 435–444.

[30] J. Ba and B. Frey, "Adaptive dropout for training deep neural networks," in *Advances in Neural Information Processing Systems*, 2013, pp. 3084–3092.

[31] B. R. Scheffers, B. F. Oliveira, I. Lamb, and D. P. Edwards, "Global wildlife trade across the tree of life," *Science (80-. ).*, vol. 366, no. 6461, pp. 71–76, Oct. 2019.

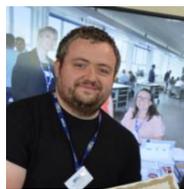

**Dr Carl Chalmers** is a Senior Lecturer in the Department of Computer Science at Liverpool John Moores University. Dr Chalmers's main research interests include the advanced metering infrastructure, smart technologies, ambient assistive living, machine learning, high performance computing, cloud computing and data visualisation. His current research area focuses on remote patient monitoring and ICT-based healthcare. He is currently leading a three-year project on smart energy data and dementia in collaboration with Mersey Care NHS Trust. The current trail involves monitoring and modelling the behaviour of dementia patients to facilitate safe independent living. In addition, he is also working in the area of high performance computing and cloud computing to support and improve existing machine learning approaches, while facilitating application integration.

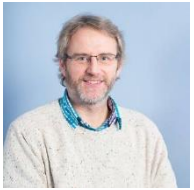

**Dr Paul Fergus** is a Reader (Associate Professor) in Machine Learning. Dr Fergus's main research interests include machine learning for detecting and predicting preterm births. He is also interested in the detection of foetal hypoxia, electroencephalogram seizure classification and bioinformatics (polygenetic obesity, Type II diabetes and multiple sclerosis). He is also currently conducting research with Mersey Care NHS Foundation Trust looking on the use of smart meters to detect activities of daily living in people living alone with Dementia by monitoring the use of home appliances to model habitual behaviors for early intervention practices and safe independent living at home.

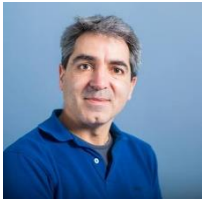

**Professor Serge** Wich is a professor in primate biology. His research focuses on on primate behavioral ecology, tropical rain forest ecology and conservation of primates and their habitats. Research is strongly focused on the Indonesian island of Sumatra and Borneo and uses a mixture of observational and experimental fieldwork. At present the key species Serge studies is the Sumatran orangutan where he is involved in research at various fieldsites of wild and reintroduced orangutans. Serge is also involved in island-wide surveys and analyses of orangutan distribution and density and the impact of land use changes on their populations. Together with Dr. Lian Pin Koh he founded ConservationDrones.org and uses drones for conservation applications.

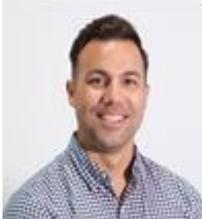

**Dr Casimiro Curbelo Montañez** is a Research Assistant in the Department of Computer Science at Liverpool John Moores University (LJMU), UK. Dr Curbelo received his B.Eng. in Telecommunications in 2011 from Alfonso X el Sabio University, Madrid (Spain). In 2014, he obtained an MSc in Wireless and Mobile Computing from LJMU. He completed his PhD in 2019 also at LJMU. He is currently part of a research team, working on strategies to detect anomalies utilising the smart metering infrastructure. His research interests include various aspects of data science, machine learning and their applications in Bioinformatics.